\documentclass[runningheads]{llncs}

\usepackage{cite}
\usepackage{amsmath,amssymb,amsfonts}
\usepackage{algorithmic}
\usepackage{graphicx}
\usepackage{textcomp}
\usepackage{xcolor}

\usepackage{wrapfig}
\usepackage{graphics} 
\usepackage{tabularx}
\usepackage{epsfig} 
\usepackage{mathptmx} 
\usepackage{amsmath} 
\usepackage{amssymb}  
\usepackage{multirow}
\usepackage{verbatim} 
\usepackage{subcaption}
\usepackage{hyperref}
\usepackage{url}
\usepackage{wrapfig}

\def\BibTeX{{\rm B\kern-.05em{\sc i\kern-.025em b}\kern-.08em
    T\kern-.1667em\lower.7ex\hbox{E}\kern-.125emX}}
\begin{document}

\title{A Study on Dense and Sparse (Visual) Rewards in Robot Policy Learning}

\author{Abdalkarim Mohtasib\inst{1} \and
Gerhard Neumann\inst{2} \and
Heriberto Cuay\'ahuitl\inst{1}}
%
\authorrunning{A. Mohtasib et al.}
%
\institute{Lincoln Centre for Autonomous Systems, University of Lincoln, Lincoln, UK 
\and
Autonomous Learning Robots, Karlsruhe Institute of Technology, Karlsruhe, Germany\\
}
\maketitle              

\begin{abstract}
Deep Reinforcement Learning (DRL) is a promising approach for teaching robots new behaviour. However, one of its main limitations is the need for carefully hand-coded reward signals by an expert. We argue that it is crucial to automate the reward learning process so that new skills can be taught to robots by their users. To address such automation, we consider task success classifiers using visual observations to estimate the rewards in terms of task success. In this work, we study the performance of multiple state-of-the-art deep reinforcement learning algorithms under different types of reward: Dense, Sparse, Visual Dense, and Visual Sparse rewards. Our experiments in various simulation tasks (Pendulum, Reacher, Pusher, and Fetch Reach) show that while DRL agents can learn successful behaviours using visual rewards when the goal targets are distinguishable, their performance may decrease if the task goal is not clearly visible. Our results also show that visual dense rewards are more successful than visual sparse rewards and that there is no single best algorithm for all tasks.  


\end{abstract}

\section{INTRODUCTION}

In Deep Reinforcement Learning, the reward signal is typically carefully designed such that the agent can learn behaviour that achieves a good performance. But hand-coding and engineering rewards requires an expert to design it for each task to be learned, and it is often not easy to design rewards for robotic tasks. 
This limits the applications of DRL to real robots, especially when the end-user of the robot has to teach the robot new tasks. To address this limitation, it is crucial to find a mechanism that can autonomously and intuitively learn the rewards from a human expert for new tasks. 

\begin{wrapfigure}{r}{0.6\textwidth}
\centerline{\includegraphics[scale=0.31]{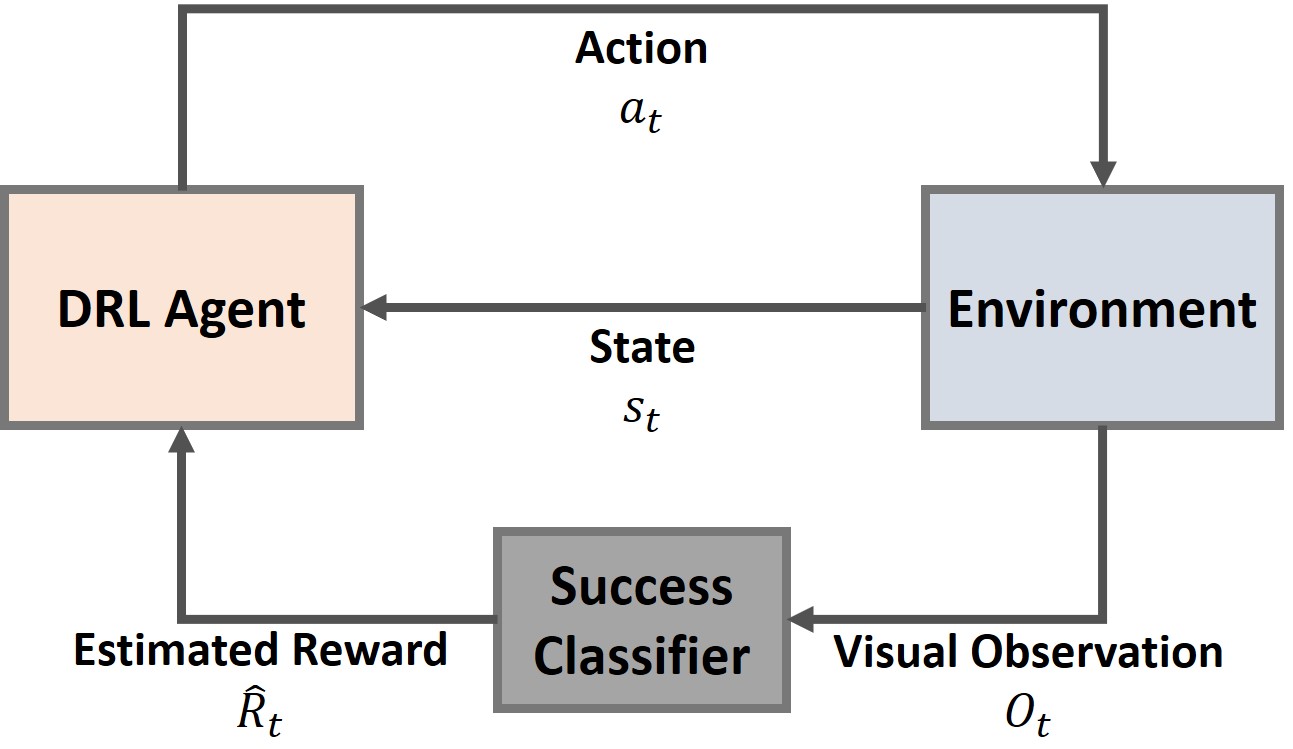}}
\caption{System Overview.}
\label{overview}
 \vspace{-5mm}
\end{wrapfigure}

The problem of autonomous reward generation has been recently investigated in the literature by several researchers. Most previous works have used image-based success classifiers---as illustrated in Fig.~\ref{overview}---to learn the task's reward \cite{vecerik2019practical,singh2019end,xie2018few,levine2018learning,ShelhamerMAD17,wang2018no,jaderberg2016reinforcement,tung2018reward,sermanet2016unsupervised}. \cite{sermanet2016unsupervised} attempted to use transfer learning to learn the rewards for new tasks, but with slow prediction times ($>0.5s$ per interaction) that prevent its practical application. 
Other approaches used goal images to estimate the reward for each time step based on the difference between the goal and the current image, calculated in different ways \cite{nair2020contextual,edwards2017cross,schoettler2019deep,sampedro2018image,edwards2016perceptual,nair2018visual}. While these approaches achieved good results in learning the task reward, they have not investigated the effects of different types of rewards on DRL agents. There is no clear study that shows how the different DRL algorithms perform with different types of reward in different tasks.

The reward learning pipeline starts with collecting expert demonstrations for the task at hand. Their images are then labelled as  success/no-success. Subsequently, the labelled data is used to train an image-based success classifier that estimates the success probability for each environment state. This success probability is used as a dense or sparse (visual) reward signal, see Section~\ref{subsec:rewards}.

The contribution of this paper is a comparison of different types of rewards (Dense, Sparse, Visual Dense, and Visual Sparse) for learning manipulation tasks. 
Our study was carried out using four different DRL algorithms (DDPG, TD3, SAC, and PPO) in four different robotic tasks. Our results show that it is indeed possible to learn good policies using visual rewards, where the higher the quality of the success classifier the better the learnt policy.
Our results also show that, while a DRL algorithm may perform very well in one task, it may perform poorly in another.

\section{Related Work}


The literature shows different ways to learn numerical rewards. Some previous works have used Inverse RL to estimate the reward function from demonstrations \cite{abbeel2004apprenticeship,finn2016guided,boularias2011relative,wulfmeier2016watch,fu2018variational}. Here, we consider a setting where the expert labels the visual observations as success/no-success. We use these expert labels to train a success classifier to estimate the reward. This setting differs from the Inverse RL setting (no expert labels). 
Other approaches use visual representations of the goal state to define the vision-based task \cite{nair2020contextual,edwards2017cross,schoettler2019deep,sampedro2018image,edwards2016perceptual,nair2018visual}. In these approaches, the goal image has been used in different ways to calculate the rewards: (i) using the latent distance between the current state image and the goal image \cite{nair2020contextual,edwards2017cross,nair2018visual}; (ii) using the pixel-wise L1 distance to the goal image \cite{schoettler2019deep}; or (iii) using the histogram distance to the goal image \cite{edwards2016perceptual}. The approach of using a goal image to represent the task's success achieved good results. Yet this approach is limited as the goal could have different varieties and shapes. Furthermore, it is not always possible to represent the task goal using one or several images. 

Our study focuses on the use of success classifiers to learn visual rewards of the task at hand \cite{vecerik2019practical,singh2019end,xie2018few,levine2018learning,ShelhamerMAD17,wang2018no,jaderberg2016reinforcement,tung2018reward,sermanet2016unsupervised}. The main neural network architecture that has been used for the task success classification is based on multiple convolutional blocks (convolutional layers followed by a max-pooling layer) followed by a multiple fully-connected layers \cite{vecerik2019practical,singh2019end,xie2018few,levine2018learning,ShelhamerMAD17,wang2018no}. Sermanet et al. \cite{sermanet2016unsupervised} used transfer learning of the Inception network \cite{szegedy2016rethinking} pre-trained for ImageNet classification \cite{deng2009imagenet} to extract the features from the environment's visual states. Subsequently, they used a simple neural network with multiple fully-connected layers to generate rewards from the extracted visual features \cite{sermanet2016unsupervised}. 
However, the interaction of such a large image classifier slows down the execution of the manipulation task. 

While some of the previous works have used dense rewards in their experiments \cite{singh2019end,levine2018learning,wang2018no,tung2018reward,sermanet2016unsupervised}, some others have only employed sparse rewards \cite{vecerik2019practical,xie2018few,ShelhamerMAD17,jaderberg2016reinforcement}. The difference between dense and sparse rewards is important because, in many tasks, the only available reward is a sparse reward and this represents a big challenge for the DRL agent to learn the task's objective. Furthermore, different types of RL algorithms have been used in these works such as DDPG \cite{vecerik2019practical}, SAC \cite{singh2019end}, A3C \cite{ShelhamerMAD17,jaderberg2016reinforcement}, REINFORCE \cite{wang2018no}, and DQN \cite{tung2018reward}. 
While task success classifiers have been used in different ways with different RL algorithms in the literature, there is no ablation study in the literature studying the pros and cons (or effects) of different types of rewards for inducing robot policies. This paper aims to fill that gap. Our ablation study, using different DRL algorithms across multiple tasks, reveals the effects of oracle dense rewards, oracle sparse rewards, visual dense rewards, and visual sparse rewards.

\section{Research Methods}


\subsection{Problem Formulation}

We consider environments that can be framed as a Markov Decision Process (MDP) \cite{bellman1957markovian}, where an agent receives a reward $r_t$ after taking action $a_t$ in the state $s_t$, then it progresses to the next state \(s_{t+1}\). We focus on the discounted case, where 
The agent tries to maximise the cumulative discounted reward  \(G_t=\mathbb{E}_\tau\left[\sum_{t=0}^{T}\gamma^t R_t\right]=\mathbb{E}_\tau\left[\sum_{t=0}^{T}\gamma^tr\left(s_t,a_t\right)\right]\), where \(\gamma\) is the discount factor, \(\tau=\left(s_0,a_0,\cdots\right)\) denotes the whole trajectory, \(s_0\sim p_0\left(s_0\right)\), \(a_t\sim\pi\left(a_t\middle|s_t\right)\), and \(s_{t+1}\sim p\left(s_{t+1}\middle|s_t,a_t\right)\).
We consider a success classifier \({\hat{R}}_t=f\left(o_t\right)\), where $o_t$ is a visual observation of the environment (an image), and \({\hat{R}}_t\in\left[0,1\right]\) is the probability of having achieved the task in state $s_t$. We train $f(o_t)$ for a new manipulation task from $N$ demonstrations by updating the parameters of this function to minimize \(\sum{\mathcal{L}(f(o_i),y_i)}\), where \(\mathcal{L}\) is the classification loss (cross entropy loss and mean square error in our case) and $y_i$ is the image label. 
We assume that a demonstrator classifies the ground truth images, which are used by  such a probabilistic classifier to learn to generate rewards.  
The research question that our study aims to answer is: {\it Can DRL agents learn good policies by using visual rewards derived from task success classifiers?}

\subsection{Rewards}
\label{subsec:rewards}

For each task, we trained DRL agents using four different types of rewards in order to  understand the effects of the different types. The agents were trained using true Dense and Sparse rewards, where they come directly from the physical simulator. The equations of Dense and Sparse rewards are shown in Table~\ref{rewards}. In addition, we used Visual Dense and Visual Sparse rewards, which were calculated based on the estimated success probability using our (best) CNN-based success classifiers. While the Visual Dense rewards for all tasks were estimated according to ${\hat{R}}_t=2\times P\left(\textrm{\it{success}}=1\middle|o_t\right)-1$, 
the Visual Sparse rewards were estimated according to ${\hat{R}}_t=
    \begin{cases}
      0, &  P\left(\textrm{\it{success}}=1\middle|o_t\right)\geq0.5\\
      -1, &  P\left(\textrm{\it{success}}=1\middle|o_t\right)<0.5\\
    \end{cases}$


\noindent Where $P\left(\textrm{\it{success}}=1\middle|o_t\right)$ is the success probability estimated by the success classifier.

\begin{table}[t]
\vspace{0.2cm}
\caption{Rewards for training DRL agents. $\phi$ is the tilt angle of the pendulum in $radians$, $D_R$ is the distance between the end-effector of the robotic arm and target position in the Reacher task, $D_P$ is the distance between the object and target location in the Pusher task, and $D_F$ is the distance between the gripper of the Fetch arm and target position.}
\begin{center}
\begin{tabular}{|c|c|c|c|c|}
\hline
{\bf Reward} & {\bf Pendulum} & {\bf Reacher} & {\bf Pusher} & {\bf Fetch} \\ 
\hline
\hline
{\bf Dense} & \(-\left|\phi\right|\)  & \(-D_R\)   &   \(-D_P\) & \(-D_F\) \\
\hline
{\bf Sparse} & 
$ 
    \begin{cases}
      0, &  \left|\phi\right|<0.15\\
      -1, &  \left|\phi\right|\geq0.15\\
    \end{cases} 
$   &
$ 
    \begin{cases}
      0, &  D_R\geq0.01m\\
      -1, &  D_R<0.01m\\
    \end{cases} 
$     &
$ 
    \begin{cases}
      0, &  D_P\geq0.01m\\
      -1, &  D_P<0.01m\\
    \end{cases} 
$     &
$ 
    \begin{cases}
      0, &  D_F\geq0.01m\\
      -1, &  D_F<0.01m\\
    \end{cases} 
$     
\\ 
\hline
\end{tabular}
\label{rewards}
\end{center}
\end{table}

\subsection{Task Success Classifiers}

\begin{figure*}[t]
    \centerline{\includegraphics[scale=0.425]{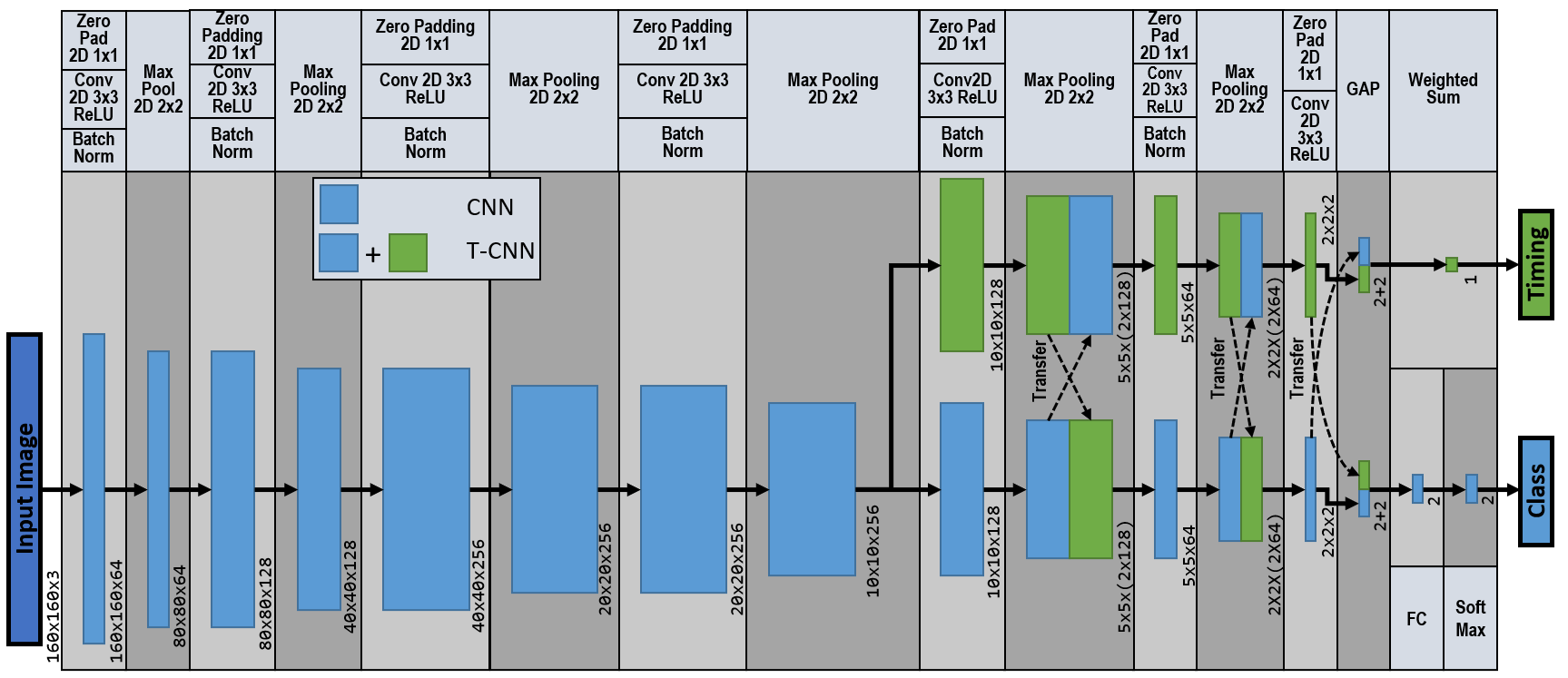}}
    \caption{Model architectures for task success classification with input images of (160 × 160 × 3). The Class output is the predicted success probability for both models. On the other hand, the Timing output is associated only with the T-CNN model. This output is the estimated task completion proportion (notation: \textbf{GAP}=Global Average Pooling).}
    \label{models}
\end{figure*}

We compare two different image classifiers trained using expert demonstrations, and use them to reward the DRL agents. 
The image classifiers are as follows. 
\begin{itemize}
\item {\bf CNN Classifier ({\bf CNN})} This is a standard CNN-based model that has been used in literature  \cite{singh2019end,vecerik2019practical,tung2018reward,levine2018learning,fu2018variational,xie2018few,edwards2017cross,sermanet2016unsupervised}. Its inputs are (160 × 160 × 3) resized images of the robotic environment, followed by six main convolutional blocks and one convolutional layer, see Fig.~\ref{models}.
\item {\bf Time-Based CNN Classifier ({\bf T-CNN})} This architecture extends the CNN one with two pathways and features (shared in between): one is the classification path, the other is a timing path that predicts the proportion of task completion (a regressor), see Fig.~\ref{models}. The task completion proportion for each image is calculated according to $y_t=\frac{t}{(j-1)}$, where $t$ is a given time step, and $j$ is the total number of time steps in the demonstration at hand. The timing path will add more gradient information and this aims to be helpful in predicting the task success.
\end{itemize}

\subsection{Training Methodology}
\label{subsec:trainmethod}

For each task, we collected a set of 10 successful demonstrations in different tasks (see Section~\ref{subsec:tasks}). These demonstrations are used for training the success classifiers in each task. Each image in these demonstrations is labeled as success/no-success. We compare the performance of the classifiers across all tasks and use the best classifier to estimate the success probabilities from visual observations. Thereafter, we train  DRL agents using four different learning algorithms\footnote{We used a PyTorch implementation of the DRL algorithms \cite{stable-baselines3}.} (DDPG \cite{lillicrap2016continuous}, TD3 \cite{fujimoto2018addressing}, SAC \cite{haarnoja2018soft}, and PPO \cite{schulman2017proximal}) with dense rewards and sparse rewards across four different tasks. Similarly, another group of DRL agents are trained but using visual dense rewards.

\section{Experiments and Results}


\subsection{Training Tasks}
\label{subsec:tasks}

\begin{figure}[b]
\vspace{0.1cm}
    \centerline{\includegraphics[scale=0.55]{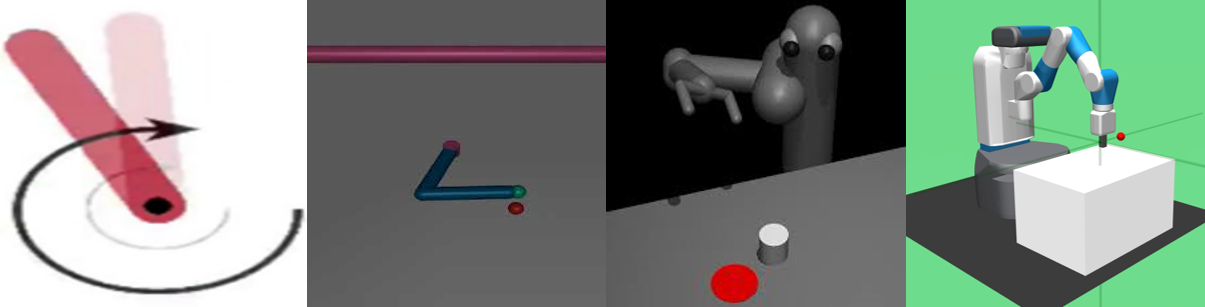}}
    \caption{Visualisation of our simulation tasks: Pendulum, Reacher, Pusher, Fetch (Reach).}
    \label{envs}
\end{figure}

We trained the DRL agents using the following OpenAI Gym Environments \cite{gym_envs}, see Fig.~\ref{envs}: (1) {{\bf Pendulum}}. A simple one Degree-Of-Freedom (DOF) task with one continuous action to stabilize the inverted pendulum in the up position. In each episode, the pendulum initial tilt angle is random. (2) {{\bf Reacher}}: In this task, the end-effector (the green point, see Fig.~\ref{envs}) of the two links robotic arm (2-DOFs) should reach the red target. The position of the red target is initialised randomly in each episode. (3) {{\bf Pusher}}: The 7-DOFs robotic arm in Fig.~\ref{envs} pushes the white object to the red target position. The position of the white object is initialised randomly in each episode. (4) {{\bf Fetch (Reach)}}: The 7-DOFs Fetch robotic arm in Fig.~\ref{envs} should reach the red target position that is initialised randomly in each episode. This is a realistic robotic task that simulates the real Fetch robot (\url{https://fetchrobotics.com}).


%
%
%
%
%

\subsection{Success Classifiers Results}
\label{subsec:classifiersresults}

The CNN (Standard Convolutional Neural Net) and T-CNN (Time-Based CNN) image classifiers in each of the four tasks were trained with a set of 10 demonstration episodes and tested with another set of 10 demonstration episodes. Here, we test the ability of the success classifier to predict the success probability for each observation (image) in the test set. We assess the performance of success classification according to the following metrics: Classification Accuracy, Precision, Recall, F1-score, Area Under the Curve. Table~\ref{results_tab1} shows the test results of our classifiers, where the T-CNN classifier outperformed the CNN classifier in all classification metrics across all tasks. This suggests that the additional gradient information for predicting the task completion proportion helps in predicting the task success. Thus, the T-CNN model is adopted to estimate the success probabilities for the visual rewards.

\begin{table}[t]
\caption{Performance results of the CNN and T-CNN classifiers (notation: \textbf{ACC}=Average Classification Accuracy, \textbf{AUC}=Area Under the Curve).}
    \begin{center}
    \begin{tabular}{|c|c|c|c|c|c||c|c|c|c|c|}
    \hline
    & \multicolumn{5}{c||}{CNN} & \multicolumn{5}{c|}{T-CNN}\\ \cline{2-11}
    \textbf{Task}  &  \textbf{ACC} & \textbf{Precision} & \textbf{Recall} & \textbf{F1 Score} & \textbf{AUC} & \textbf{ACC} & \textbf{Precision} & \textbf{Recall} & \textbf{F1 Score} & \textbf{AUC} \\
    \hline \hline
    \textbf{Pendulum}     & 1.000 & 1.000 & 1.000 & 1.000 & 1.000  & 1.000 & 1.000 & 1.000 & 1.000 & 1.000 \\
    \textbf{Reacher}      & 0.738 & 0.970 & 0.704 & 0.816 & 0.962 & 0.872 & 0.970 & 0.872 & 0.918 & 0.982 \\
    \textbf{Pusher}       & 0.990 & 0.992 & 0.994 & 0.993 & 1.000 & 0.992 & 0.994 & 0.994 & 0.994 & 1.000 \\
    \textbf{Fetch}        & 0.898 & 0.908 & 0.982 & 0.943 & 0.976 & 0.948 & 0.966 & 0.975 & 0.970 & 0.989 \\ \hline
    \textbf{Average}      & 0.907 & 0.968 & 0.920 & 0.938 & 0.985 & \textbf{0.953} & \textbf{0.990} & \textbf{0.960} & \textbf{0.971} & \textbf{0.993} \\ \hline
    %
    \end{tabular}
    \label{results_tab1}
    \end{center}
    \end{table}
    
\subsection{Experimental Results of the DRL Agents}

We evaluated different aspects in the performance of our DRL agents. First, we start with the learning curves of the DRL agents under the different settings as shown in Table~\ref{results_tab2}. The most important outcome from these learning curves is that the DRL agents (except for PPO agents) were able to learn good policies by using only the visual rewards that come from the success classifier. See the following video for example behaviours of the trained DRL agents\footnote{Video: \href{https://youtu.be/8zOqEQDBleU}{https://youtu.be/8zOqEQDBleU}}.

\begin{table*}[!t]

\caption{Learning curves of DRL agents using different learning algorithms (DDPG, TD3, SAC, PPO) across four tasks when trained using dense, sparse, visual dense, and visual sparse rewards. The agents used five different seeds, 320 learning curves in total.}
\begin{center}

\begin{tabular}{|c|c|c|c|c|}
\hline
\textbf{}     & \textbf{Dense Reward} & \textbf{Sparse Reward} & \textbf{Visual Dense Reward} & \textbf{Visual Sparse Reward} \\ \hline
\multirow{6}{*}{\rotatebox[origin=c]{90}{\textbf{Pendulum}}} & \multirow{6}{*}{\includegraphics[scale=0.108]{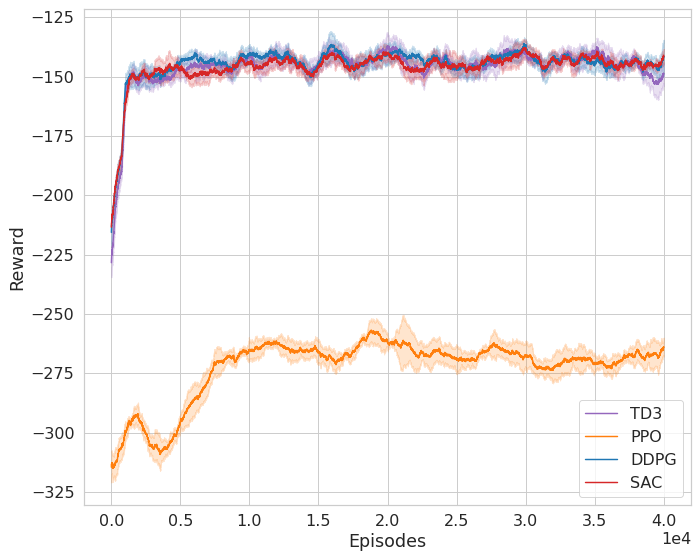}}  &   \multirow{6}{*}{\includegraphics[scale=0.108]{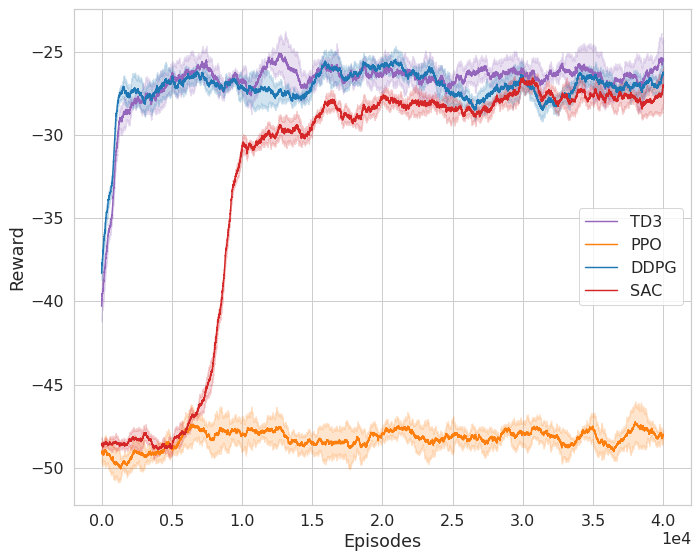}} & \multirow{6}{*}{\includegraphics[scale=0.108]{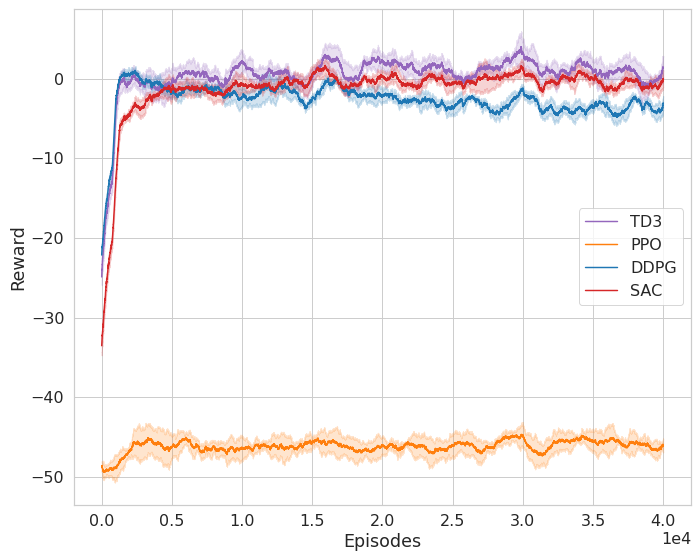}} & \multirow{6}{*}{\includegraphics[scale=0.108]{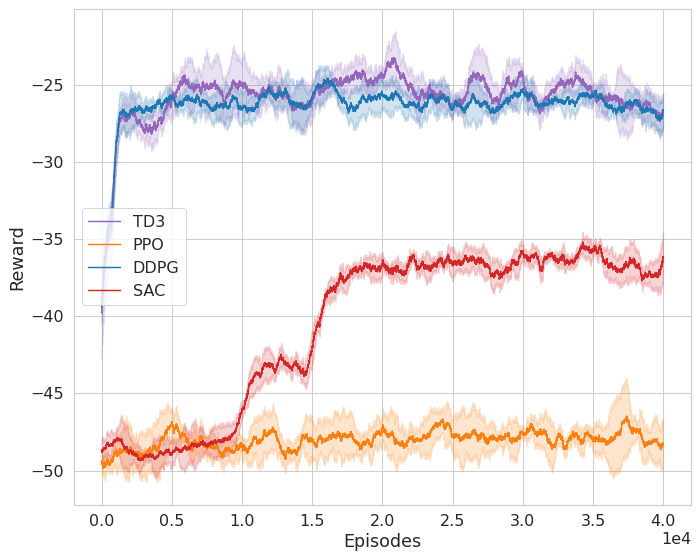}} \\

 & & & & \\  & & & & \\  & & & & \\  & & & & \\  & & & & \\ \hline
 
\multirow{6}{*}{\rotatebox[origin=c]{90}{\textbf{Reacher}}} & \multirow{6}{*}{\includegraphics[scale=0.108]{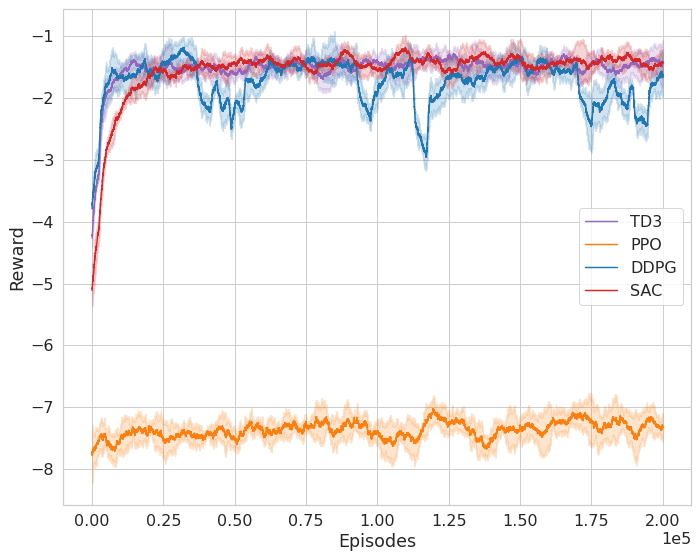}}  &   \multirow{6}{*}{\includegraphics[scale=0.108]{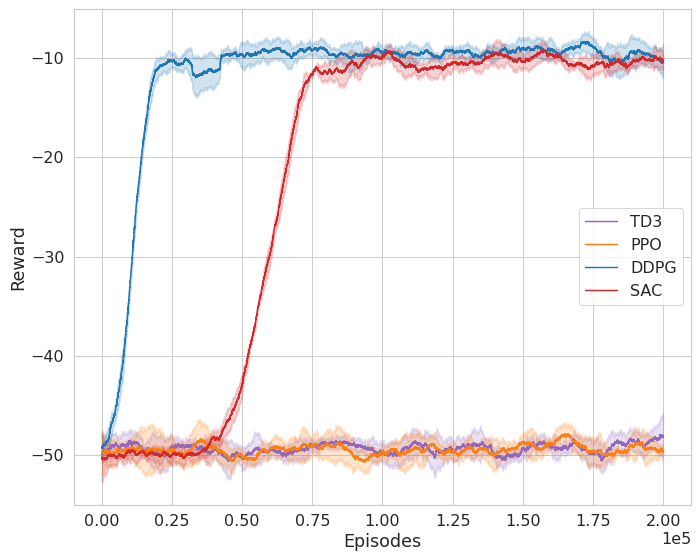}} & \multirow{6}{*}{\includegraphics[scale=0.108]{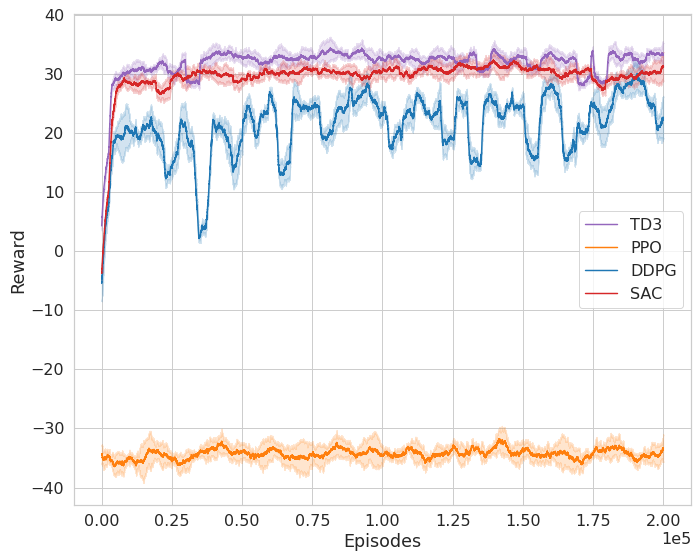}} & \multirow{6}{*}{\includegraphics[scale=0.108]{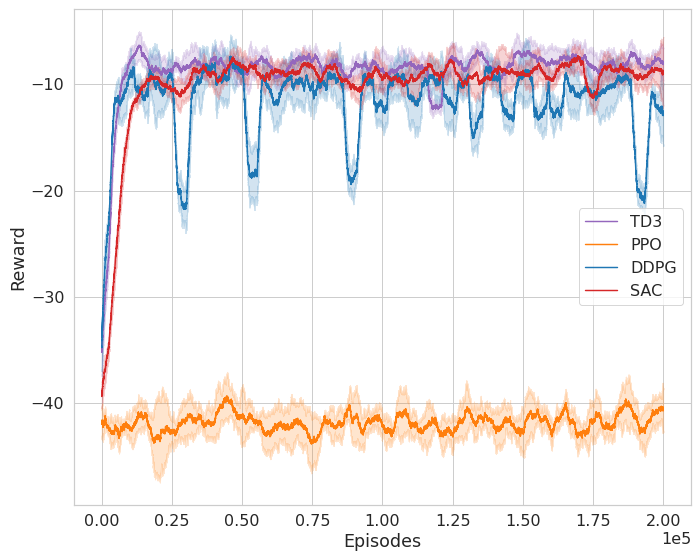}} \\

 & & & & \\  & & & & \\  & & & & \\  & & & & \\  & & & & \\ \hline

\multirow{6}{*}{\rotatebox[origin=c]{90}{\textbf{Pusher}}} & \multirow{6}{*}{\includegraphics[scale=0.108]{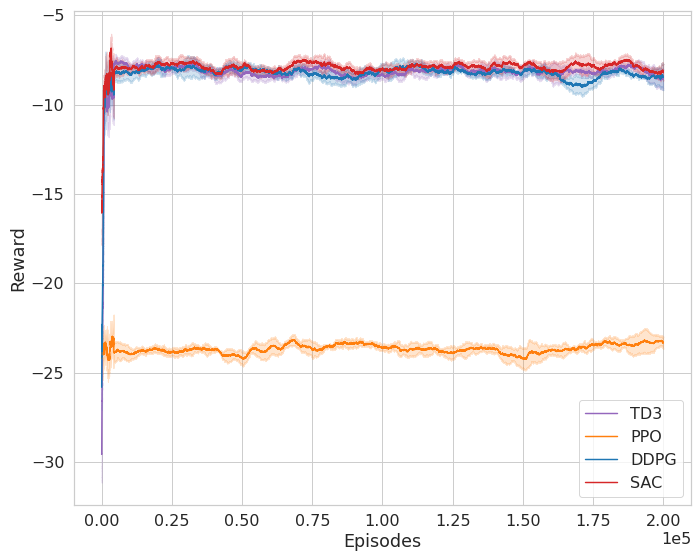}}  &   \multirow{6}{*}{\includegraphics[scale=0.108]{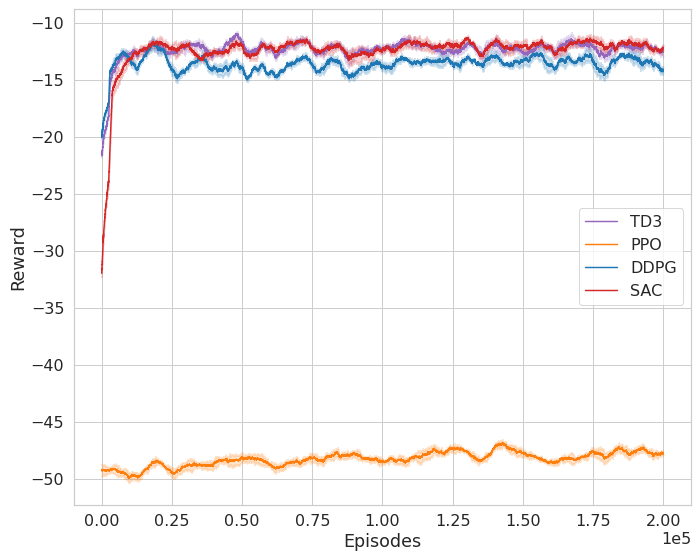}} & \multirow{6}{*}{\includegraphics[scale=0.108]{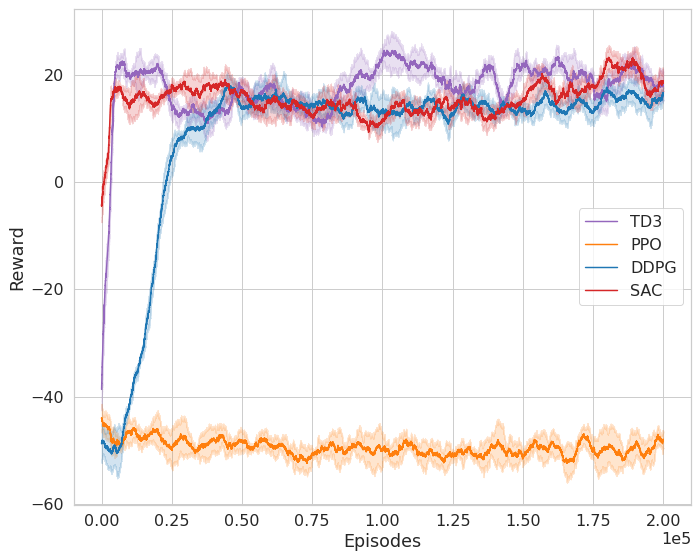}} & \multirow{6}{*}{\includegraphics[scale=0.108]{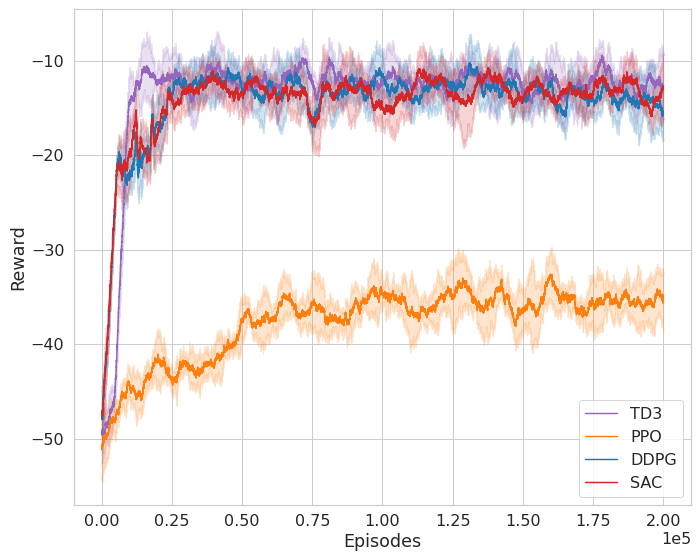}} \\

 & & & & \\  & & & & \\  & & & & \\  & & & & \\  & & & & \\ \hline

\multirow{6}{*}{\rotatebox[origin=c]{90}{\textbf{Fetch (Reach)}}} & \multirow{6}{*}{\includegraphics[scale=0.108]{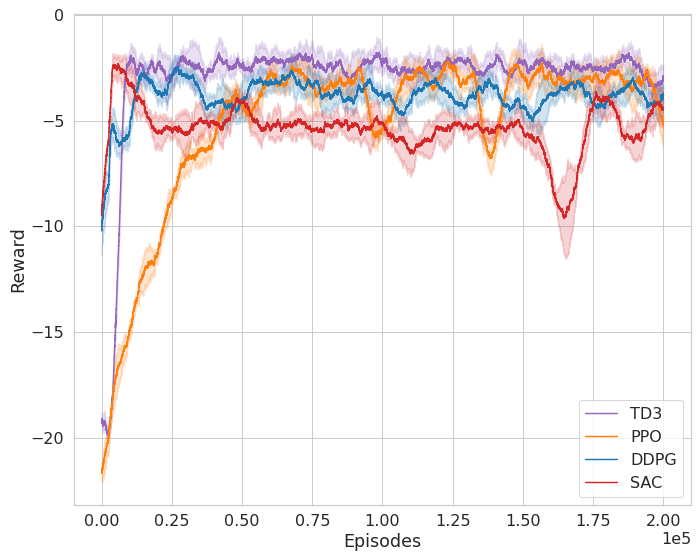}}  &   \multirow{6}{*}{\includegraphics[scale=0.108]{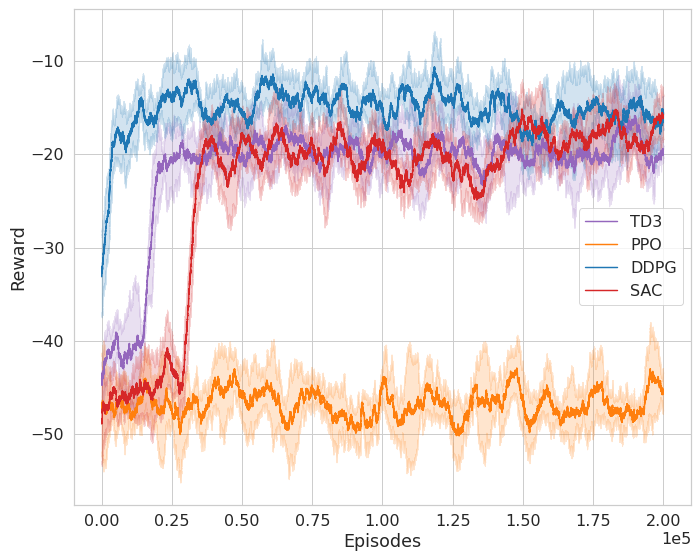}} & \multirow{6}{*}{\includegraphics[scale=0.108]{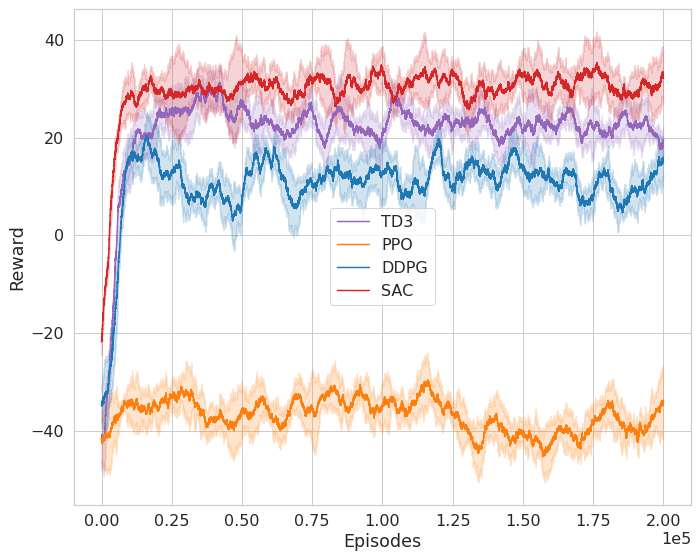}} & \multirow{6}{*}{\includegraphics[scale=0.108]{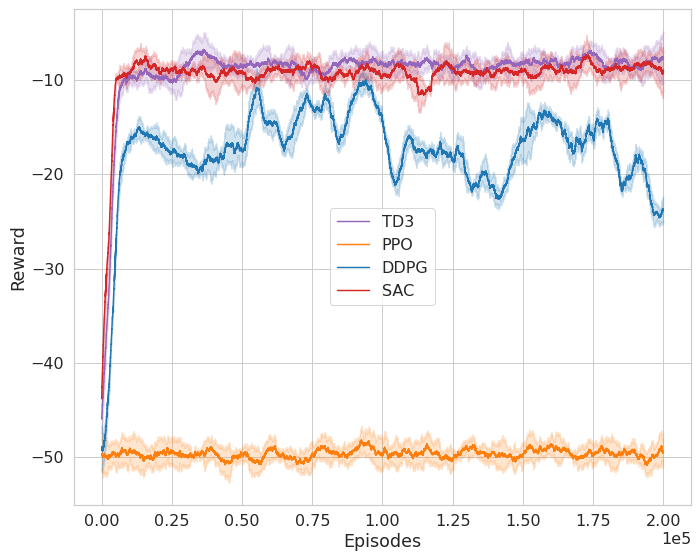}} \\

 & & & & \\  & & & & \\  & & & & \\  & & & & \\  & & & & \\ \hline
 
\end{tabular}
\label{results_tab2}
\end{center}
\end{table*}

It is crucial to test the learned policies to ensure that the visual rewards can be used to learn useful behaviours that lead to the successful execution of the tasks. Table~\ref{results_tab3} shows the test results of the learned policies. It is clear from these results that the visual rewards are indeed helpful in learning good successful behaviours---as noted by their success rates across tasks.  On average, there is a small drop in performance when using visual rewards as the success classifier is error-prone. It can be noted that the drop in performance when using visual rewards is larger in the Reacher and the Fetch (Reach) tasks. With further investigation and experiments, we found that when the target object is behind the robotic arm, the visual images are not reflecting the correct environment's state. Thus, the success classifier fails to predict the correct success probability, and hence the drop in the performance in this task. The ranking of algorithms according to average success rate is as follows:  DDPG (81.4\%$\pm$18\%), SAC (77.8\%$\pm$20\%), TD3 (75.7\%$\pm$26\%), and PPO (12.1\%$\pm$21\%).

\begin{table}[!ht]
\caption{Performance of DRL algorithms across tasks. The maximum episode's length is 100 steps. Training and test times of DRL agents: training in HH:MM, and test in seconds. While 2 million steps were used in the {\bf  Pendulum} task, 10 million steps were used in the other tasks. The learnt policies were tested with 1000 episodes in each task.}
\begin{center}
\small
\begin{tabular}{|c|c|cc|cc|cc|cc|}
\hline
\multirow{3}{*}{\textbf{Task}}     & \multirow{3}{*}{\textbf{Agent}} & \multicolumn{2}{c|}{\textbf{Dense Reward}} & \multicolumn{2}{c|}{\textbf{Sparse Reward}} & \multicolumn{2}{c|}{\textbf{Visual Dense Rew.}} & \multicolumn{2}{c|}{\textbf{Visual Sparse Rew.}} \\ \cline{3-10} 

    &  & \textbf{Success}  & \textbf{Average}  & \textbf{Success}   & \textbf{Average}  & \textbf{Success}      & \textbf{Average}     & \textbf{Success}      & \textbf{Average}    \\ 
    
    &  & \textbf{Rate} & \textbf{Eps. Len.}  & \textbf{Rate} & \textbf{Eps. Len.}  & \textbf{Rate} & \textbf{Eps. Len.}     & \textbf{Rate} & \textbf{Eps. Len.}    \\ \hline \hline
    
\multirow{4}{*}{\rotatebox[origin=c]{90}{\textbf{Pendulum}}} & \textbf{DDPG}  & 98\%  & 62.36 & 96\% & 59.68 & 96\% & 63.18 & 87\% & 53.88 \\ 

    & \textbf{TD3}  & 100\% & 62.53 & 87\% & 60.56 & 99\% & 58.19 & 87\% & 63.9  \\
                                   
    & \textbf{SAC}  & 95\%  & 66.37 & 75\% & 61.76 & 99\% & 62.7  & 38\% & 61.12 \\ 
                                   
    & \textbf{PPO}  & 1\%   & 60.77 & 2\%  & 61.19 & 2\%  & 51.86 & 4\%  & 61.66 \\ \hline 
                                   
\multirow{4}{*}{\rotatebox[origin=c]{90}{\textbf{Reacher}}}   & \textbf{DDPG}  & 100\% & 51.87 & 97\% & 53.79 & 45\% & 82.16 & 62\% & 75.35 \\ 

                                    & \textbf{TD3}  & 98\%  & 54.64 & 8\%  & 97.28 & 52\% & 78.34 & 33\% & 83.99 \\
                                   
                                    & \textbf{SAC}  & 98\%  & 52.93 & 97\% & 56.09 & 45\% & 80.38 & 47\% & 73.95 \\ 
                                   
                                    & \textbf{PPO}  & 6\%   & 98.13 & 8\%  & 95.01 & 11\% & 95.12 & 10\% & 95.6  \\ \hline 
                                   
\multirow{4}{*}{\rotatebox[origin=c]{90}{\textbf{Pusher}}}   & \textbf{DDPG}  & 91\%  & 60.44 & 90\% & 59.94 & 80\% & 60.52 & 72\% & 69.38 \\

                                   & \textbf{TD3}   & 92\%  & 62.8  & 87\% & 67.95 & 83\% & 66.25 & 80\% & 69.89 \\ 
                                   
                                   & \textbf{SAC}   & 95\%  & 58.17 & 90\% & 63.16 & 87\% & 67.21 & 84\% & 63.1  \\ 
                                   
                                   & \textbf{PPO}   & 2\%   & 99.46 & 19\% & 91.78 & 4\%  & 98.78 & 14\% & 95.27 \\ \hline
                                   
\multirow{4}{*}{\rotatebox[origin=c]{90}{\textbf{Fetch}}}   & \textbf{DDPG}  & 91\%  & 52.23 & 88\% & 60.95 & 58\% & 73.25 & 51\% & 74.78 \\

                                   & \textbf{TD3}   & 94\%  & 53.87 & 82\% & 58.07 & 64\% & 67.53 & 65\% & 68.13 \\ 
                                   
                                   & \textbf{SAC}   & 81\%  & 58.52 & 82\% & 62.5  & 72\% & 67.83 & 59\% & 75.29 \\ 
                                   
                                   & \textbf{PPO}   & 90\%  & 52.75 & 4\%  & 98.11 & 13\% & 92.95 & 4\%  & 97.41 \\ \hline 
\multicolumn{2}{|c|}{{Avg. Success}} & 77.00\% & 62.99 & 63.25\% & 69.24 & 56.88\% & 72.89 & 49.81\% & 73.92 \\ \hline
\multicolumn{2}{|c|}{{Std. Success}} & 37.02\% & 14.67 & 38.89\% & 16.03 & 34.28\% & 13.85 & 29.78\% & 13.13 \\ \hline
\end{tabular}
\label{results_tab3}
\end{center}
\vskip -0.3in
\end{table}


A statistical analysis using the Wilcoxon Signed-Rank Test (paired)\cite{wilcoxon1992individual} on the results of Table~\ref{results_tab3} revealed the following. Comparing Dense Success Vs. Sparse Success, the p-values are: $p=1e-4$ including PPO, and $p=6e-7$ excluding PPO. Comparing Visual Dense Success Vs. Visual Sparse Success, the p-values are: $p=4e-4$ including PPO, and $p$=0.016 excluding PPO. Whilst the first comparison supports our claim that dense rewards are better than sparse ones, the second supports the claim that visual dense rewards are better than visual sparse rewards.

We carried out another statistical analysis using the Wilcoxon Signed-Rank Test (paired)\cite{wilcoxon1992individual} to compare the ranking of  algorithms according to average success rate. While comparing DDPG Vs. TD3 gives a  $p$-value of 0.236, comparing DDPG Vs. SAC gives $p=$0.182. The differences are not significant and more comparisons are needed. 


Table~\ref{results_tab5} reports training and test times of various experimented settings. Considering the training time\footnote{PC: \textbf{CPU}: Intel i7-6950 @ 3.00GHz, 10 cores. \textbf{RAM}: 32GB. \textbf{GPU}: NVIDIA TITAN X 12GB.} of agents using visual rewards, their training time is almost twice than non-visual rewards. 
Although such long training times should be addressed in future work, this cost comes with a large benefit where there is no need to hand-code the reward functions. Similarly and in contrast to agents using non-visual rewards, the test times of agents using visual rewards increase by about 30ms for every environment step. We calculated the average test time for one environment step across tasks, which resulted in $\sim$45ms---acceptable for real robotic tasks. 

Furthermore, we investigated the effects of the choice of success classifier by comparing policies using our baseline and proposed success classifiers---CNN and T-CNN, respectively. Results show that while the performance of agents is similar in simple tasks (74\% of task success on avg. across algorithms for both classifiers in the Pendulum task), T-CNN-based agents outperform CNN-based agents in more complex tasks (61.7\% and 47.8\% of an overall average task success across tasks, respectively), suggesting that the higher the performance of success classifiers the better learnt policies.

\begin{table}[!t]
\caption{Training and test times of DRL agents: training in HH:MM, and test in seconds. While 2 million steps were used in the {\bf  Pendulum} task, 10 million steps were used in the other tasks. The learnt policies were tested with 1000 episodes in each task.}
\begin{center}
\begin{tabular}{|c|c|cccc|cccc|}
\hline
 & \multirow{2}{*}{\textbf{Task}} & \multicolumn{4}{c|}{\textbf{Dense or Sparse Reward}}  & \multicolumn{4}{c|}{\textbf{Visual Reward }} \\
\cline{3-10} 
 &  & \textbf{DDPG} & \textbf{TD3} & \textbf{SAC} & \textbf{PPO} & \textbf{DDPG}   & \textbf{TD3}   & \textbf{SAC}   & \textbf{PPO}   \\ \hline \hline
\multirow{5}{*}{\rotatebox[origin=c]{0}{\bf{Training}}} 
& Pendulum          & 05:45 & 05:47 & 09:10 & 04:12 & 16:22 & 16:15 & 22:27 & 10:30 \\ 
& Reacher           & 27:02 & 19:18 & 45:50 & 12:16 & 45:24 & 51:42 & 77:56 & 18:40 \\ 
& Pusher            & 29:07 & 27:49 & 35:01 & 12:11 & 47:00 & 45:59 & 60:35 & 29:58 \\ 
& Fetch (Reach)     & 29:30 & 24:33 & 34:50 & 14:59 & 50:20 & 40:25 & 62:55 & 26:34 \\ \cline{2-10} 
& Avg.              & 22:51 & 19:21 & 31:12 & 10:54 & 39:46 & 38:35 & 55:58 & 21:25 \\ \hline \hline
\multirow{5}{*}{\rotatebox[origin=c]{00}{\textbf{Test}}} 
 & Pendulum      & 17.03 & 17.28 & 16.22 & 16.75 & 46.02 & 33.15 & 33.90 & 43.35 \\
 & Reacher       & 16.53 & 17.42 & 17.75 & 17.18 & 45.60 & 43.55 & 47.63 & 49.03 \\
 & Pusher        & 17.54 & 17.09 & 15.84 & 16.76 & 44.53 & 45.17 & 47.93 & 46.59 \\
 & Fetch (Reach) & 16.14 & 17.56 & 17.38 & 17.78 & 45.72 & 45.34 & 45.87 & 47.13 \\ \cline{2-10}
 & Avg.          & 16.81 & 17.34 & 16.80 & 17.12 & 45.47 & 41.80 & 43.83 & 46.53 \\
\hline
\end{tabular}
\label{results_tab5}
\end{center}
\vskip -0.2in
\end{table}

\section{Conclusion and Future Work}
This paper shows that it is indeed possible to learn successful policies from visual rewards, though with higher computational cost than non-visual rewards. Our experiments reveal the following. First, dense rewards can achieve higher task success than sparse rewards. Second, the better the success classifier the better the policy. Third, when images do not represent the correct state of the environment, this may lead to learning poor policies. Fourth, while one algorithm might be good in a given task, it may not performs well in another task. DDPG achieved the highest task success across tasks, but the differences in performance against other algorithms were not significant.

Future work will consist of investigating the proposed learned visual rewards on real robotic tasks and multiple robot platforms. Other future works with high potential contribution to the previous work include accelerating the training times of DRL agents, and improving their success rates across tasks.
    
{
\scriptsize
\bibliographystyle{splncs04}
\bibliography{mybib}

\begin{thebibliography}{10}
\providecommand{\url}[1]{\texttt{#1}}
\providecommand{\urlprefix}{URL }
\providecommand{\doi}[1]{https://doi.org/#1}

\bibitem{abbeel2004apprenticeship}
Abbeel, P., Ng, A.Y.: Apprenticeship learning via inverse reinforcement
  learning. In: ICML (2004)

\bibitem{bellman1957markovian}
Bellman, R.: A markovian decision process. Journal of mathematics and mechanics
   \textbf{6}(5) (1957)

\bibitem{boularias2011relative}
Boularias, A., Kober, J., Peters, J.: Relative entropy inverse reinforcement
  learning. In: AISTATS (2011)

\bibitem{gym_envs}
Brockman, G., Cheung, V., Pettersson, L., Schneider, J., Schulman, J., Tang,
  J., Zaremba, W.: Openai gym (2016)

\bibitem{deng2009imagenet}
Deng, J., Dong, W., Socher, R., Li, L.J., Li, K., Fei-Fei, L.: Imagenet: A
  large-scale hierarchical image database. In: CVPR (2009)

\bibitem{edwards2016perceptual}
Edwards, A., Isbell, C., Takanishi, A.: Perceptual reward functions. arXiv
  preprint arXiv:1608.03824  (2016)

\bibitem{edwards2017cross}
Edwards, A.D., Sood, S., Isbell~Jr, C.L.: Cross-domain perceptual reward
  functions. arXiv preprint arXiv:1705.09045  (2017)

\bibitem{finn2016guided}
Finn, C., Levine, S., Abbeel, P.: Guided cost learning: Deep inverse optimal
  control via policy optimization. In: ICML (2016)

\bibitem{fu2018variational}
Fu, J., Singh, A., Ghosh, D., Yang, L., Levine, S.: Variational inverse control
  with events: A general framework for data-driven reward definition. NIPS
  \textbf{31} (2018)

\bibitem{fujimoto2018addressing}
Fujimoto, S., Hoof, H., Meger, D.: Addressing function approximation error in
  actor-critic methods. In: ICML (2018)

\bibitem{haarnoja2018soft}
Haarnoja, T., Zhou, A., Abbeel, P., Levine, S.: Soft actor-critic: Off-policy
  maximum entropy deep reinforcement learning with a stochastic actor. In: ICML
  (2018)

\bibitem{jaderberg2016reinforcement}
Jaderberg, M., Mnih, V., Czarnecki, W.M., Schaul, T., Leibo, J.Z., Silver, D.,
  Kavukcuoglu, K.: Reinforcement learning with unsupervised auxiliary tasks.
  arXiv preprint arXiv:1611.05397  (2016)

\bibitem{levine2018learning}
Levine, S., Pastor, P., Krizhevsky, A., Ibarz, J., Quillen, D.: Learning
  hand-eye coordination for robotic grasping with deep learning and large-scale
  data collection. IJRR  \textbf{37}(4-5) (2018)

\bibitem{lillicrap2016continuous}
Lillicrap, T.P., Hunt, J.J., Pritzel, A., Heess, N., Erez, T., Tassa, Y.,
  Silver, D., Wierstra, D.: Continuous control with deep reinforcement
  learning. In: ICLR (2016)

\bibitem{nair2020contextual}
Nair, A., Bahl, S., Khazatsky, A., Pong, V., Berseth, G., Levine, S.:
  Contextual imagined goals for self-supervised robotic learning. In: CoRL
  (2020)

\bibitem{nair2018visual}
Nair, A.V., Pong, V., Dalal, M., Bahl, S., Lin, S., Levine, S.: Visual
  reinforcement learning with imagined goals. NIPS  \textbf{31} (2018)

\bibitem{stable-baselines3}
Raffin, A., Hill, A., Ernestus, M., Gleave, A., Kanervisto, A., Dormann, N.:
  Stable baselines3. \url{https://github.com/DLR-RM/stable-baselines3} (2019)

\bibitem{sampedro2018image}
Sampedro, C., Rodriguez-Ramos, A., Gil, I., Mejias, L., Campoy, P.: Image-based
  visual servoing controller for multirotor aerial robots using deep
  reinforcement learning. In: IROS (2018)

\bibitem{schoettler2019deep}
Schoettler, G., Nair, A., Luo, J., Bahl, S., Ojea, J.A., Solowjow, E., Levine,
  S.: Deep reinforcement learning for industrial insertion tasks with visual
  inputs and natural rewards. arXiv preprint arXiv:1906.05841  (2019)

\bibitem{schulman2017proximal}
Schulman, J., Wolski, F., Dhariwal, P., Radford, A., Klimov, O.: Proximal
  policy optimization algorithms. arXiv preprint arXiv:1707.06347  (2017)

\bibitem{sermanet2016unsupervised}
Sermanet, P., Xu, K., Levine, S.: Unsupervised perceptual rewards for imitation
  learning. arXiv preprint arXiv:1612.06699  (2016)

\bibitem{ShelhamerMAD17}
Shelhamer, E., Mahmoudieh, P., Argus, M., Darrell, T.: Loss is its own reward:
  Self-supervision for reinforcement learning. In: ICLR (2017)

\bibitem{singh2019end}
Singh, A., Yang, L., Hartikainen, K., Finn, C., Levine, S.: End-to-end robotic
  reinforcement learning without reward engineering. In: RSS (2019)

\bibitem{szegedy2016rethinking}
Szegedy, C., Vanhoucke, V., Ioffe, S., Shlens, J., Wojna, Z.: Rethinking the
  inception architecture for computer vision. In: CVPR (2016)

\bibitem{tung2018reward}
Tung, H.Y., Harley, A., Huang, L.K., Fragkiadaki, K.: Reward learning from
  narrated demonstrations. In: CVPR (2018)

\bibitem{vecerik2019practical}
Vecerik, M., Sushkov, O., Barker, D., Roth{\"o}rl, T., Hester, T., Scholz, J.:
  A practical approach to insertion with variable socket position using deep
  reinforcement learning. In: ICRA (2019)

\bibitem{wang2018no}
Wang, X., Chen, W., Wang, Y.F., Wang, W.Y.: No metrics are perfect: Adversarial
  reward learning for visual storytelling. In: ACL (2018)

\bibitem{wilcoxon1992individual}
Wilcoxon, F.: Individual comparisons by ranking methods. In: Breakthroughs in
  statistics, pp. 196--202. Springer (1992)

\bibitem{wulfmeier2016watch}
Wulfmeier, M., Wang, D.Z., Posner, I.: Watch this: Scalable cost-function
  learning for path planning in urban environments. In: IROS (2016)

\bibitem{xie2018few}
Xie, A., Singh, A., Levine, S., Finn, C.: Few-shot goal inference for
  visuomotor learning and planning. In: CoRL (2018)

\end{thebibliography}
}
\end{document}